\documentclass[10pt, conference, compsocconf]{IEEEtran}
\ifCLASSOPTIONcompsoc
  \usepackage[nocompress]{cite}
\else
  \usepackage{cite}
\fi
\ifCLASSINFOpdf
\else
\fi
\usepackage{xspace}
\usepackage{graphicx}
\usepackage{float}
\usepackage{epstopdf}
\usepackage{color}
\usepackage{float}
\usepackage{amsmath}
\usepackage{algorithm}
\usepackage{algorithmic}

\begin{document}
%
\title{SenseGen: A Deep Learning Architecture for Synthetic Sensor Data Generation}

\author{\IEEEauthorblockN{Moustafa Alzantot}
\IEEEauthorblockA{University of California, Los Angeles\\
malzantot@ucla.edu}
\and
\IEEEauthorblockN{Supriyo Chakraborty}
\IEEEauthorblockA{IBM T. J. Watson Research Center\\
supriyo@us.ibm.com}
\and
\IEEEauthorblockN{Mani Srivastava}
\IEEEauthorblockA{University of California, Los Angeles\\
mbs@ucla.edu}}


%


\maketitle

\begin{abstract}
Our ability to synthesize sensory data that preserves specific statistical properties of the real data has had tremendous implications on data privacy and big data analytics. The synthetic data can be used as a substitute for selective real data segments -- that are sensitive to the user -- thus protecting privacy and resulting in improved analytics. However, increasingly adversarial roles taken by data recipients such as mobile apps, or other cloud-based analytics services, mandate that the synthetic data, in addition to preserving statistical properties, should also be ``difficult’’ to distinguish from the real data. Typically, visual inspection has been used as a test to distinguish between datasets. But more recently, sophisticated classifier models (discriminators), corresponding to a set of events, have also been employed to distinguish between synthesized and real data. The model operates on both datasets and the respective event outputs are compared for consistency.

Prior work on data synthesis have often focussed on classifiers that are built for features explicitly preserved by the synthetic data. This suggests that an adversary can build classifiers that can exploit a potentially disjoint set of features for differentiating between the two datasets. In this paper, we take a step towards generating sensory data that can pass a deep learning based discriminator model test, and make two specific contributions: first, we present a deep learning based architecture for synthesizing sensory data. This architecture comprises of a \emph{generator model}, which is a stack of multiple Long-Short-Term-Memory (LSTM) networks and a Mixture Density Network (MDN); second, we use another LSTM network based \emph{discriminator model} for distinguishing between the true and the synthesized data. Using a dataset of accelerometer traces, collected using smart-phones of users doing their daily activities, we show that the deep learning based discriminator model can only distinguish between the real and synthesized traces with an accuracy in the neighborhood of 50\%.
\end{abstract}


%
\IEEEpeerreviewmaketitle

\section{Introduction}

A large number of data recipients (e.g., mobile apps, and other cloud-based big-data analytics) rely on the collection of personal sensory data from devices such as smartphones, wearables, and home IoT devices to provide services such as remote health monitoring~\cite{diaz2015fitbit}, location tracking~\cite{wang2012no}, automatic indoor map construction and navigation~\cite{alzantot2012crowdinside} and so on. However, the prospect of sharing sensitive personal data, often  prohibits large-scale user adoption and therefore the success of such systems. To circumvent these issues and increase data sharing, synthetic data generation has been used as an alternative to real data sharing. The generated data preserves only the required statistics of the real data (used by the apps to provide service) and nothing else and are used as a substitute for selective real data segments — that are sensitive to the user —  thus protecting privacy and resulting in improved analytics.

However, increasingly adversarial roles taken by the data recipients mandate that the synthetic data, in addition to preserving statistical properties, should also be ``difficult'' to distinguish from the real data. Even in non-adversarial settings, analytics services can behave in unexpected ways if the input data is different from the expected data, thereby requiring the synthesized and real datasets to exhibit ``similarity''. Typically, visual inspection has been used as a test to distinguish between datasets. But more recently, sophisticated classifier models (discriminators), corresponding to a set of events, have also been employed to distinguish between synthesized and real data. The model operates on both datasets and the respective event outputs are compared for consistency. In fact, prior work on data synthesis have often focussed on classifiers that are built for features explicitly preserved by the synthetic data. This suggests that an adversary can build classifiers that can exploit a potentially disjoint set of features for differentiating between the two datasets.

In this paper, we present SenseGen -- a deep learning based \emph{generative model} for synthesizing sensory data. While deep learning methods are known to be capable of generating realistic data samples, training them was considered to be difficult requiring large amounts of data. However, recent work on generative models such as \emph{Generative Adversarial Networks}~\cite{goodfellow2014generative, goodfellow2016nips} (GAN) and variational auto-encoders ~\cite{kingma2013auto, salimans2016improved} have shown that it is possible to train these models with moderate sized datasets. GANs have proven successful in generating different types of data including photo-realistic high resolution images~\cite{ledig2016photo}, realistic images from text description~\cite{reed2016generative}, and even for new text and music composition~\cite{yu2016seqgan},~\cite{bowman2015generating}. Furthermore, inspired by the architecture of GANs, we also use a deep learning based discriminator model. The goal of the generator model is to synthesize data that can pass the discriminator test that is designed to distinguish between synthesized and real data. Note, unlike prior work on data synthesis, a deep learning based discriminator is not trained on a pre-determined set of features. Instead, it continuously learns the best set of features that can be used to differentiate between the real and synthesized data — making it hard for the generator to pass the discriminator test.

To summarize, we make two contributions. First, we present a deep learning based architecture for synthesizing sensory data. This architecture comprises of a \emph{generator model}, which is a stack of multiple Long-Short-Term-Memory (LSTM) networks and a Mixture Density Network (MDN). Second, we use another LSTM network based \emph{discriminator model} for distinguishing between the true and the synthesized data. Using a dataset of accelerometer traces, collected using smart-phones of users doing their daily activities, we show that the deep learning based discriminator model can only distinguish between the real and synthesized traces with an accuracy in the neighborhood of 50\%.

The rest of this paper is organized as follows: 
Section~\ref{sec:model} provides a description for our model architecture and the training algorithm used. This is followed by Section~\ref{sec:result} that describes our experimental design and initial results. Finally, Section~\ref{sec:conc} concludes the paper.


\section{Model Design}
\label{sec:model}

Sensors data, e.g. accelerometer, gyroscope, barometer, etc., are represented as a sequence of values $ \mathbf{x} = (\mathbf{x}_{1}, \mathbf{x}_{2}, ..., \mathbf{x}_{T})$ where $\mathbf{x}_{i} \in \mathbf{R}^{d}$, for $i = 1,\ldots,|T|$ where $d$ is the dimensionality of the time series (i.e. $d=3$ in case of 3-axis accelerometer ) and $T$ is the number of time steps for which the data has been collected.

SenseGen consists of two deep learning models:
\begin{itemize}
\item Generator ($\mathcal{G}$): The generator $\mathcal{G}$ is capable of generating new synthetic time series data from random noise input.

\item Discriminator ($\mathcal{D}$): The goal of the discriminator $\mathcal{D}$ is to assess the quality of the examples generated by the generator $\mathcal{G}$.
\end{itemize}

Both $\mathcal{G}$ and $\mathcal{D}$ are based on recurrent neural network models which have shown a lot of success in sequential data modeling. We describe the model details below.

\begin{algorithm}
\caption{Training algorithm}
\label{alg:training}
\begin{algorithmic}[1]
\FOR{$t=1,2,\ldots, T$}
\STATE Sample $\mathcal{X}_{true}$ minibatch from true data
\STATE Sample $\mathcal{X}_{gen}$ minibatch from the generative model $G$
\STATE Train the discriminative model $\mathcal{D}$ on the training set $(\mathcal{X}_{true},\mathcal{X}_{gen})$ for 200 epochs

\STATE Sample another $\mathcal{X}_{true}$ minibatch from true data
\STATE Sample another $\mathcal{X}_{gen}$ minibatch from the generative model $G$
\STATE Train the generative model $\mathcal{G}$ on the training set $(\mathcal{X}_{true})$ for 100 epochs
\ENDFOR
\end{algorithmic}
\end{algorithm}

\subsection{Generative Model}
\label{sec:gen}

\begin{figure*}[!h]
\centering
\includegraphics[scale=0.78]{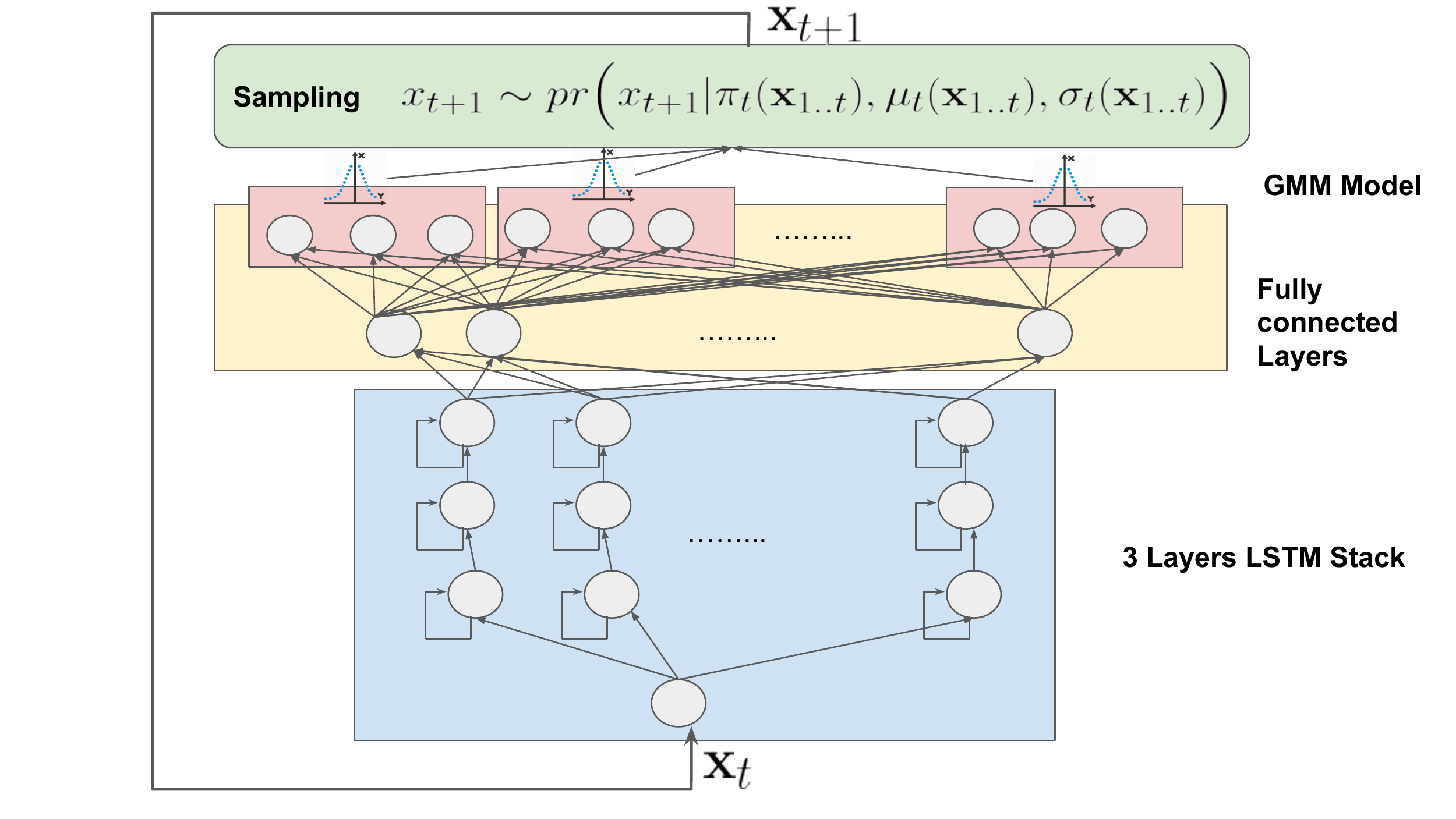} 
\label{fig:genmodel}
\caption{Generative Model architecture}
\end{figure*}

Recurrent neural networks (RNN) are a class of neural networks which are distinguished by having units with feedback cycles which allows the units to maintain a memory of state about the previous inputs. This makes them suitable for handling tasks dealing with sequential time-series inputs.  The input time-series is applied to the neural network units one step at time. Each RNN artificial neuron (often called RNN unit or RNN cell) maintains a hidden internal state memory $h_t$ which is updated at each time-step according to the new input $x_t$ and previous internal state memory value $h_{t-1}$ 

\[ h_t = \sigma (W_{hh}  h_{t-1} + W_{xh}  x_t + b_h) \]
where $\sigma(x) $ is the sigmoid activation function
\[\sigma(x) = \frac{1}{1 + e^{-x}} \] 
Also each unit generates another time-series of outputs $o_t$ as a function of the internal memory state which is computed according to the following equation:
\[ o_t = tanh (W_o  h_{t} + b_o)\]

The set $ \theta = \{W_{hh}, W_{xh}, b_h, W_o, b_o\}$ represents the RNN cell parameters. The RNN training algorithm picks the values of $\theta$ that minimizes the defined loss function.

In order to handle complex time-series sequences, Multiple RNN units can be used at the same layer and also multiple RNN units can be stacked on top of each other such that the time series of outputs from the RNN units at one layer are used as inputs to the RNN units on top of them.  This way, we can design more powerful  recurrent neural networks which are both deep and wide. Like other neural networks, we train a recurrent neural  networks  possible by using a modified version of back-propagation known as  back-propagation through time (BPTT) algorithm~\cite{werbos1990backpropagation}. However, RNN units suffer from two major problems during training deep models over long time-series inputs. First, it is the \emph{vanishing gradient} problem, where the error gradient goes to zero during propagation presenting difficulty while learning the weights of early layers or capturing long-term depedencies. Second, it is the \emph{exploding gradient} problem, where the gradient value might grow exponentially causing numerical errors in the training algorithm. These two problems present a major hurdle in training RNNs. To solve the exploding gradient problem, the gradient value is clipped at each unit, while modified architectures of RNN units such as the Long Short Term Memory (LSTM)~\cite{hochreiter1997long}  and Gated Recurrent Units (GRU)~\cite{chung2015gated} have been introduced to come over the vanishing gradient problem. 

LSTM units are modified version of the standard RNN units that add three additional gates inside the RNN unit : input gate $(i_t)$, forget gate $(f_t)$ and output gate $(o_t)$. The values of these gates are computed as functions of the unit's internal cell state $c_{t}$ and current input $x_t$. These gates are used to control what information being stored in the unit's internal memory $h_t$ to avoid vanishing gradient problem and become better in remembering sequence dependencies for longer range.  The gates, internal memory and LSTM unit output at each time step are computed according to the following equations:

\[f_t = \sigma (W_{xf} x_t + W_{hf} h_{t-1} + b_f ) \]
\[i_t = \sigma (W_{xi} x_t + W_{hi} h_{t-1} + b_i ) \]
\[o_t = \sigma (W_{xo} x_t + W_{ho} h_{t-1} + b_o ) \]
\[c_t = f_t \odot c_{t-1} + i_{t} \odot tanh(W_{hc} h_{t-1} + W_{xc} x_{t} + b_c )  \]
\[h_t = o_t \odot tanh(c_t) \]
where $\odot$ is the elementwise multiplication.  In the  rest of the paper, we define the function $LSTM$ that maps the current input $x_t$ and current LSTM unit output $h_t$ to  new output as an abstraction of the previous LSTM update equations.
\[ h_t = LSTM(x_t, h_{t-1}) \]
Like the standard RNN units, LSTM units can also be stacked on top of each other in order to model complex time-series data. We use LSTMs in our model because they are successful in modeling sequences with long-term dependencies.

Recurrent Neural networks can be used for the generation of a sequence with any length by predicting the sequence one step at a time. At each time step, the network output $y_t$ is used to define a probability distribution for the next step $x_{t+1}$ value.
\[ x_{t+1} \sim pr(x_{t+1} | y_t) \]

 The value $x_{t+1}$ is then fed back into the model as a new input to predict another time step. By repeatedly  doing this, it is  theoretically possible to generate a sequence of any length. However, the choice of output distribution becomes critical and must be chosen carefully to represent the type of data we are generating. The simplest choice that we consider the output $y_t$ as the next step sample $x_{t+1} = y_t$. and then we define the loss as the root mean squared difference between the sequence of inputs and the sequence of predictions.
\[ \mathcal{L}^{\mathcal{G}}(\theta_{\mathcal{G}}) = \sum_{t=1}^{T}{(x_{t} - y_{t})^2}   \]

Then we train the whole model by using gradient descent to minimize the loss value. However, we find this setup to be incapable of generating good sensory data sequences for the following reasons:
\begin{itemize}
\item Since all RNN update equations are deterministic, this means that if you try generating sequences from a given start input value $x_0$ (usually starting by zero) the model will generate the same sequence again at every-time.
\item Assigning the model output as the next sample means that the next sample distribution is a uni-model distribution with zero variance. Because for sensory data at a given step more than one value can be a good choice for the next step a uni-modal prediction is not enough. A more flexible generation of sensory data requires probabilistic sampling from a multi-modal distribution on top of the RNN.
\end{itemize}

As a solution for these issues, we use Mixture Density Network (MDN)~\cite{bishop1994mixture}. Mixture density network is a combination of a neural network and mixture distribution. The outputs of neural network are used to specify the weights of the mixtures and the parameters of each distribution. \cite{graves2013generating} shows how MDN with Gaussian mixture model (GMM) defined on top of a recurrent neural network is successful in learning how to generate highly realistic handwriting by predicting the pen location one point at a time.

Our generative model architecture is shown in Figure \ref{fig:genmodel}. At the bottom we have a stack of 3 layers $(l^{(1)}_t , l^{(2)}_t ,l^{(3)}_t  )$ of LSTM units. Each layer has 256 units. 
\[ l^{(1)}_t = LSTM(l^{(1)}_{t-1}, x_t ) \]
\[ l^{(2)}_t = LSTM(l^{(2)}_{t-1}, l^{(1)}_t  ) \]
\[ l^{(3)}_t = LSTM(l^{(2)}_{t-1}, l^{(1)}_t  ) \]

 The output from the last LSTM layer is feed into a fully connected layer with 128 units with sigmoid activations. 
\[l^{(4)}_t = \sigma(W_4 l^{(3)}_t + b_4 ) \]
where $W_4 \in \mathbf{R}^{256x128}, b_4 \in \mathbf{R}^{128}$
 The final layer is another fully connected layer with $72$ output units.
\[l^{(5)}_t = \sigma(W_5 l^{(4)}_t + b_5 ) \]

where $W_5 \in \mathbf{R}^{128x72}, b_5 \in \mathbf{R}^{72}$. The outputs from the last layer $l^{(5)}_t$ are used as the weights and parameters of the output Gaussian mixture model (GMM).
\[ \pi_t(\mathbf{x}_{1..t}) = softmax( l^{(5)}_t{[1 ... 24]} ) \]
\[ \mu_t(\mathbf{x}_{1..t}) = l^{(5)}_t{[25 ... 48]} \]
\[ \sigma_t(\mathbf{x}_{1..t}) = e^{(l^{(5)}_t{[49 ... 72]})} \]

where the softmax function:
\[ softmax(x_k) = \frac{e^{x_k}}{\sum_{j=1}^{24}{e^{x_j}} } \]
is used to ensure that weights defined by $pi_t$ are normalized (i.e $\sum_{k=1}^{24}{\pi_k} = 1$), and the exponential function while computing the standard deviation of the Gaussians $\sigma_t$ is meant to ensure that the $\sigma_t$ is positive.
The mixture weights $\pi_t$, guassian means $\mu_t$, and standard deviations $\sigma_t$ are used to define to a probability distribution for the next output
\begin{equation*}
\begin{aligned}
pr(x_{t+1} | & \pi_t, \mu_t,\sigma_t  ) = & \\ 
&  \sum_{k=1}^{24} {\pi^{k}_t(\mathbf{x}_{1..t}) * \mathcal{N}  (x_{t+1}; \mu^{k}_t(\mathbf{x}_{1..t}), \sigma^{k}_t(\mathbf{x}_{1..t})} ) &
\end{aligned}
\end{equation*}

from which we can sample the predicted next step value.
\[x_{t+1} \sim pr \Big(x_{t+1} | \pi_t(\mathbf{x}_{1..t}), \mu_t(\mathbf{x}_{1..t}),\sigma_t(\mathbf{x}_{1..t}) \Big)  \]

The whole model is trained end-to-end by RMSProp~\cite{tieleman2012lecture} and truncated back-propagation through time with a cost function $\mathcal{L}(\theta_{\mathcal{G}})$ defined to increase the likelihood of generating the next timestep value. This is the equivalent to minimizing the negative log  likelihood $\mathcal{L}(\theta_{\mathcal{G}})$ with respect to the set of generative model parameters $\theta_{\mathcal{G}}$.

\[\mathcal{L}^{\mathcal{G}}(\theta_{\mathcal{G}}) = -\sum_{t=1}^{T}{\log \left( pr(x_{t+1} | \pi_t(\mathbf{x}_{1..t}), \mu_t(\mathbf{x}_{1..t}),\sigma_t(\mathbf{x}_{1..t})  ) \right)}  \]

\subsection{Discriminative Model}
\label{sec:discr}
In order to quantify the similarity between the generated time-series and the real sensor timeseries collected from users. We build another model $\mathcal{D}$ whose goal is to distinguish between samples generated by $\mathcal{G}$. The discriminative model $\mathcal{D}$ is trained to distinguish between the samples coming from the dataset for real sensor traces $\mathcal{X}_{true}$ and others samples from the dataset $\mathcal{X}_{gen}$ which is generated by the model $\mathcal{G}$.

The architecture of model $\mathcal{D}$ consists of a layer of 64 LSTM  units followed by a fully connected layers with 16 hidden units using sigmoid activation function and an output layer with a single unit with sigmoid activation function. The output value this discriminative model when a given an input sensor values timeseries $\mathbf{x}_{test}$ is interpreted as the probability that the given input timeseries is coming from the real  dataset $\mathcal{X}_{true}$.

\[\mathcal{D}(\mathbf{x}_{test}) = Pr(\mathbf{x}_{test} \in \mathcal{X}_{true} ) \]

We train the model $\mathcal{D}$ in a supervised way by using a training data consists from mini-batchs of $m$ samples from the real data dataset $\mathcal{X}_true$ with their target output = 1, and other mini-batchs of $m$ samples generated from the the model $\mathcal{G}$ with their target output = 0. Each samples is a time series of 400 steps. The training aims to minimize the cross-entropy loss $ \mathcal{L}^{\mathcal{D}}$ with respect to the set of discriminitive model parameters.
\[ \mathcal{L}^{\mathcal{D}}(\theta_{\mathcal{D}})= - \left( \sum_{i=1}^{m}{ \log{(\mathcal{D}(\mathcal{X}_{true}^{(i)}))} + \log{(1-\mathcal{D}(\mathcal{X}_{gen}^{(i)}))} } \right) \]


\section{Results and Analysis}
\label{sec:result}
For our experiments and evaluation studies, We use the Human Activity Recognition database~\cite{anguita2012human} as our training data. The HAR database contains accelerometer and gyroscope recordings of 30 individuals while performing activities of daily living (ADL) (Walking, walking upstairs, walking downstairs, sitting, standing, and laying). Accelerometer and gyroscope were collected at 50Hz from a Samsung Galaxy SII phone attached to the user's the waist. The accelerometer and gyroscope values were pre-processed to compute the linear acceleration (by removing the gravity component).\\
\begin{figure*}[t!]
\centering
\includegraphics[scale=0.75]{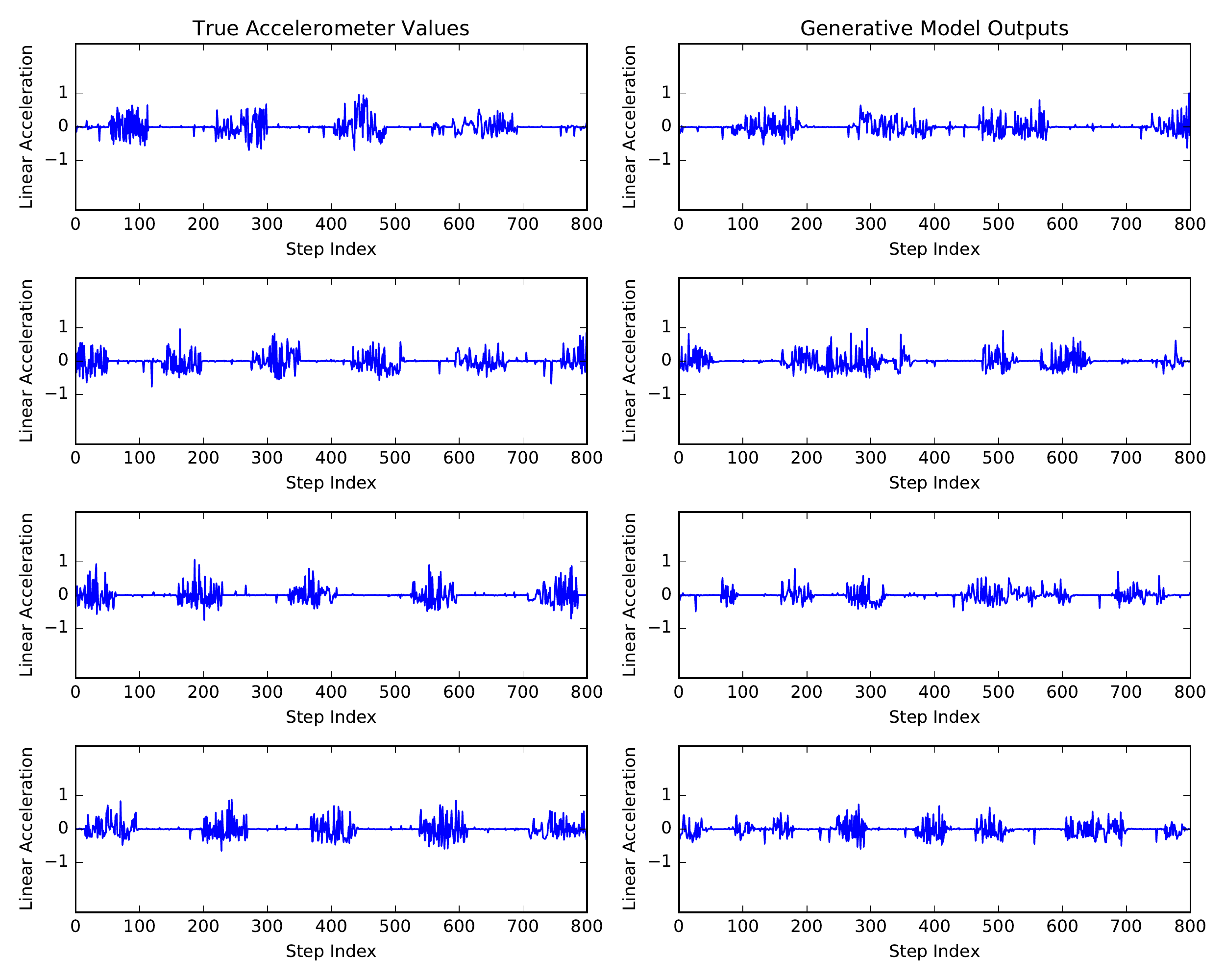}
\caption{Visual comparison between the real and generated accelerometer time-series samples}
\label{fig:vis_compare}
\end{figure*}
We train the deep learning model using Google TensorFlow~\cite{abadi2016tensorflow} deep learning framework $r0.11$ on Nvidia GTX Titan X GPU with 3,584 CUDA cores running 11 TFLOPS with 12 GB Memory.  The training takes about 5 hours until the generative model converges after 20,000 epochs when trained on a time-series of 7000 time steps.

\begin{figure}[!h]
\centering
\includegraphics[scale=0.6]{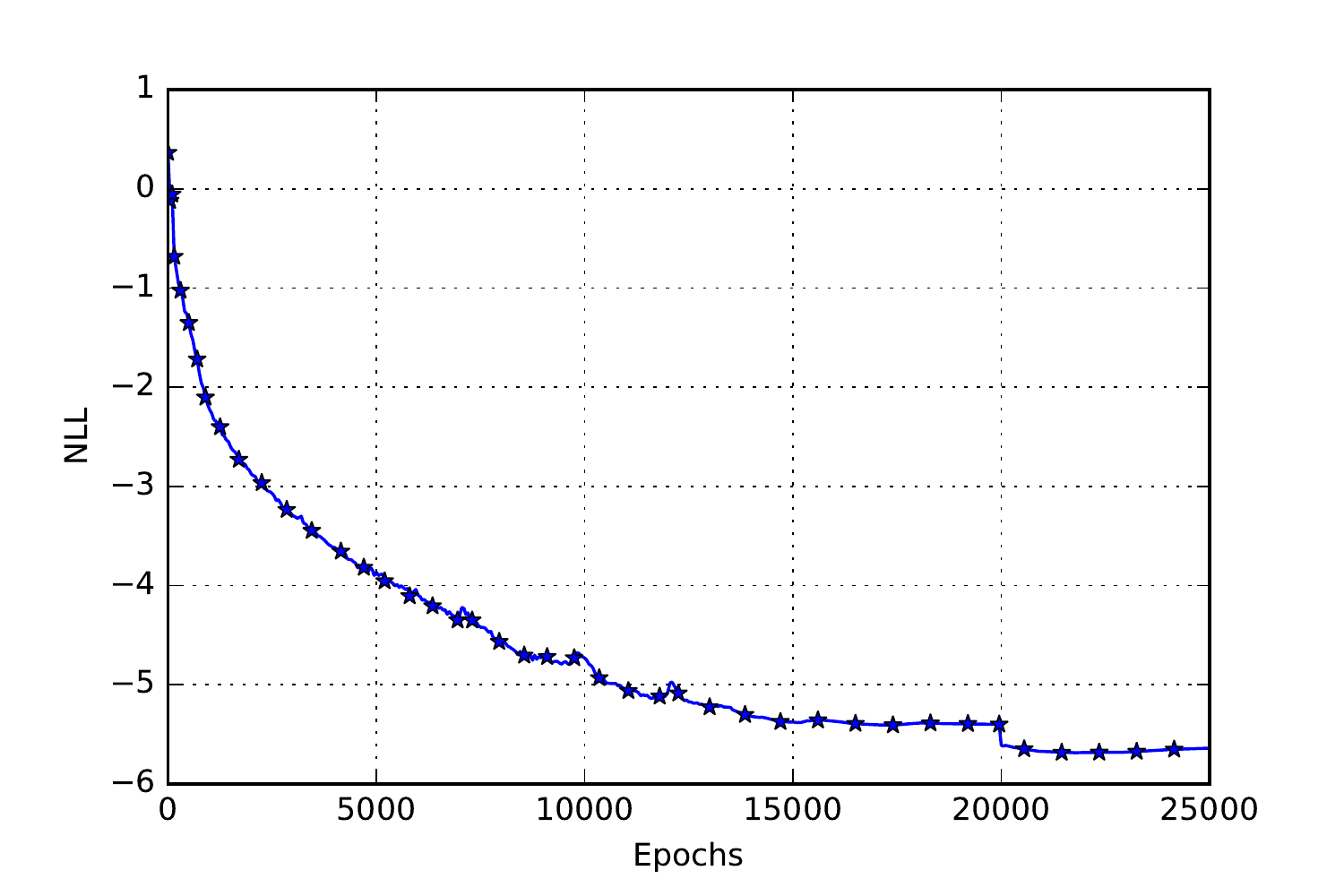}
\label{fig:gen_loss}
\caption{Negative Log likelihood cost of the generative model during training}
\end{figure}

Evaluating a generative model is challenging because it is hard to find one metric that quantifies how realistic the output looks and also how novel is it compared to the training data (to avoid the trap of having a model that just remembers the input training data and outputs it again). These metrics should be specific according to the type of the data the model is trained on. Prior work on generative models for images resort to human judgment of output samples quality.  In our work, we use the following methods for qualification:

\textbf{Generative loss during training} We show how the loss of the generative model goes down while training. 
Figure~\ref{fig:gen_loss} shows the negative log likelihood cost of the generative model $\mathcal{L}^{\mathcal{G}}$ during training.  This means that  the model is becoming  better in assigning higher probability for the true next-step values during prediction.  

\begin{figure}[!t]
\centering
\includegraphics[scale=0.3]{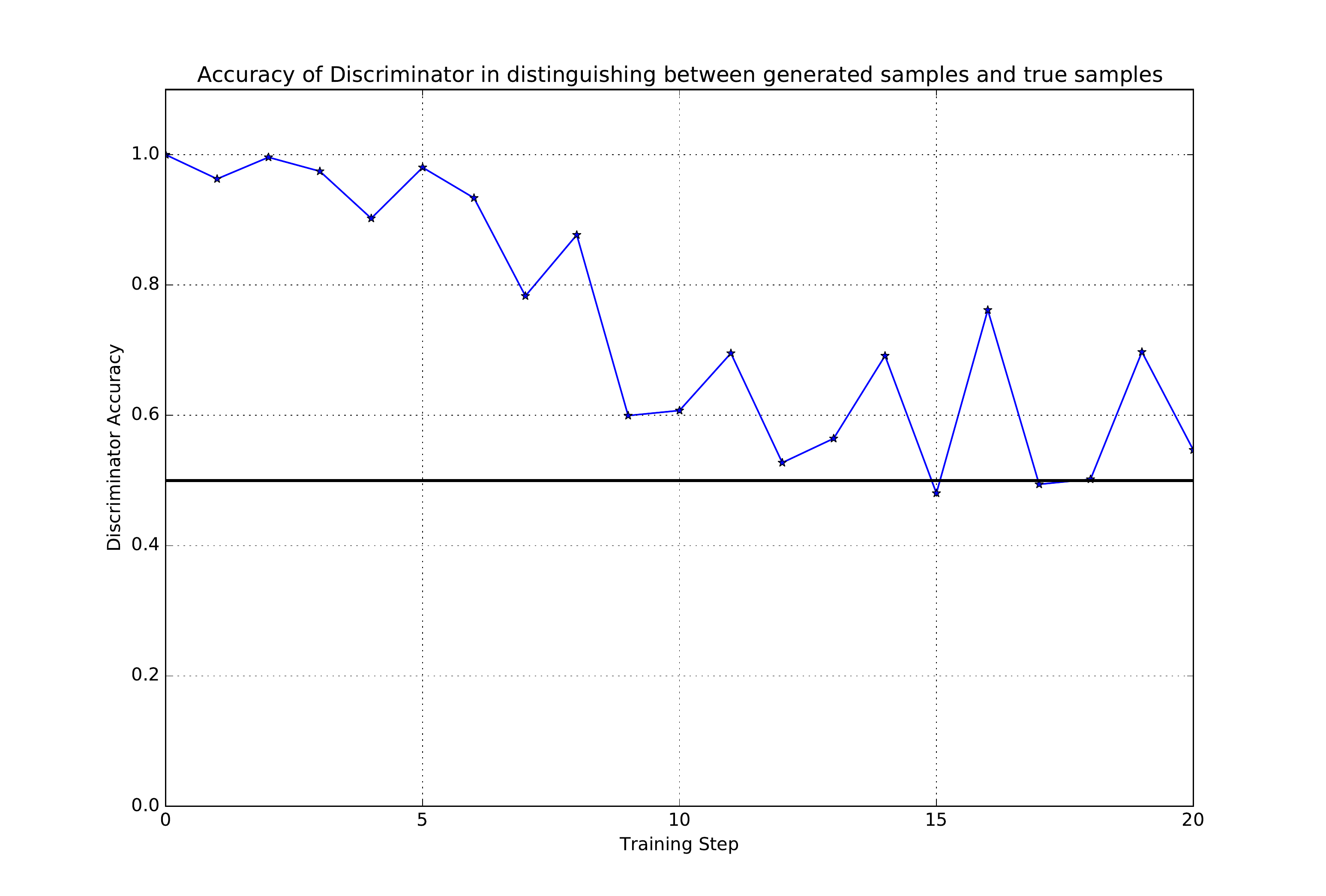}
\caption{Accuracy of Discriminator $\mathcal{D}$ in deciding which samples are not real}
\label{fig:disc_acc}
\end{figure}

\textbf{Generative loss during training} Figure~\ref{fig:vis_compare} shows a visual comparison between 4 random samples generated by generative model $\mathcal{G}$ and 4 random subset of real accelerometer time-series values from the HAR dataset.

\textbf{Indistinguishability between synthesized and real data samples} We use another deep learning model $\mathcal{D}$ whose is training  to quantify the differences between the real samples and the synthesized samples.  Figure ~\ref{fig:disc_acc} shows how the  accuracy of this  model goes down as training continues. At the beginning the  accuracy in deciding whether input samples are synthesized is almost 100\% However, as we train the models for more epochs, the accuracy of model $\mathcal{D}$ in identifying the synthetic samples reduces to around $50\%$.


\section{Conclusion}
In this paper, we outlined our initial experiences of using a deep learning based architecture for synthesizing time series of sensory data. We identified that the synthesized data should be able to pass a deep learning based discriminator test designed to distinguish between the synthesized and true data. We then demonstrated that our generator  can be successfully used to beat such a discriminator by restricting its accuracy to around $50\%$.

Our generator-discriminator model pair is a GAN-similar architecture. However, due to the difficulties of doing back-propagation through the MDN-based stochastic network, we do not yet incorporate \textit{adversarial training} by feeding back the discriminator output into the generator training. we hope to close the feedback loop between the discriminator and the generator model for synthesizing even more effective data samples.
\label{sec:conc}
\section*{Acknowledgement}
This research was sponsored by the U.S. Army Research Laboratory and the U.K. Ministry of Defence under Agreement Number W911NF-16-3-0001. The views and conclusions contained in this document are those of the authors and should not be interpreted as representing the official policies, either expressed or implied, of the U.S. Army Research Laboratory, the U.S. Government, the U.K. Ministry of Defence or the U.K. Government. The U.S. and U.K. Governments are authorized to reproduce and distribute reprints for Government purposes notwithstanding any copy-right notation hereon.






\bibliographystyle{IEEEtran}
\bibliography{sensegan_ref}

\end{document}